\documentclass[conference]{IEEEtran}
\IEEEoverridecommandlockouts

\usepackage{cite}
\usepackage{amsmath,amssymb,amsfonts}
\usepackage{algorithmic}
\usepackage{graphicx}
\usepackage{textcomp}
\usepackage{xcolor}
\usepackage{widetable}
\usepackage{hyperref}
\usepackage{subcaption}
\usepackage{comment}
\usepackage{multirow}
\usepackage{braket}
\usepackage{cleveref}
\def\BibTeX{{\rm B\kern-.05em{\sc i\kern-.025em b}\kern-.08em
    T\kern-.1667em\lower.7ex\hbox{E}\kern-.125emX}}

    \usepackage[T1]{fontenc}
\usepackage{times}
\usepackage{graphicx}
\begin{document}

\title{Multi-VQC: A Novel QML Approach for Enhancing Healthcare Classification \vspace{-10pt}}

\author{\IEEEauthorblockN{Antonio Tudisco\IEEEauthorrefmark{1},
Deborah Volpe\IEEEauthorrefmark{2}, and  Giovanna Turvani\IEEEauthorrefmark{1} }
\IEEEauthorblockA{\IEEEauthorrefmark{1}Department of Electronics and Telecommunications,
Politecnico di Torino
Italy\\
\IEEEauthorrefmark{2}Istituto Nazionale di Geofisica e Vulcanologia, Rome, 00143, Italy\\
\href{mailto:antonio.tudisco@polito.it}{antonio.tudisco@polito.it},
 \href{mailto:deborah.volpe@ingv.it}{deborah.volpe@ingv.it}, and
\href{mailto:giovanna.turvani@polito.it}{giovanna.turvani@polito.it}}
\vspace{-30pt}}

\maketitle

\begin{abstract}
Accurate and reliable diagnosis of diseases is crucial in enabling timely medical treatment and enhancing patient survival rates. In recent years, Machine Learning has revolutionized diagnostic practices by creating classification models capable of identifying diseases. However, these classification problems often suffer from significant class imbalances, which can inhibit the effectiveness of traditional models. Therefore, the interest in Quantum models has arisen,  driven by the captivating promise of overcoming the limitations of the classical counterpart thanks to their ability to express complex patterns by mapping data in a higher-dimensional computational space. 

This work proposes a novel approach for enhancing the classification performance of Quantum Neural Networks (QNN) consisting of multiple Variational Quantum Circuits (VQCs) arranged sequentially. This strategy increases the nonlinearity of the model by exploiting the measurement operation and improving its ability to capture complex patterns. In this analysis, the proposed method is compared against classical models while varying its degrees of freedom, specifically the number of involved VQCs, on three well-known healthcare datasets --- Prostate Cancer, Heart Failure, and Diabetes.
The results prove the potential of the quantum model and demonstrate the validity of the proposed approach, showing that its advantage increases with the complexity of the classification.  
\end{abstract}


\vspace{-15pt}
\section{Introduction}
Accurate and timely diagnosis of diseases such as cancer or heart failure is essential for quick medical intervention and improving patient survival rates. In recent years, \textbf{Machine Learning} (\textbf{ML}) has revolutionized diagnostic practices by developing classification models \cite{an2023comprehensive} that detect diseases using features extracted from patient data, including biomarkers, genetic profiles, and medical imaging. These models exploit algorithms such as logistic regression and support vector machines, automating pattern recognition to identify slight indicators that might be challenging for human experts to discern.
However, classification tasks in this context are often characterized by significant class imbalance, as the prevalence of the disease is much lower than that of healthy cases --- or, in some instances, vice versa --- within the training datasets. This imbalance can result in models that are biased toward the majority class, leading to poor sensitivity and low recall for the minority (disease) class. Consequently, addressing data imbalance has become a critical focus for improving model reliability \cite{haixiang2017learning}.

Quantum models offer an attractive alternative to classical ML techniques \cite{10398184}, harnessing their ability to detect complex and intricate patterns by operating in a higher-dimensional computational space enabled by the principles of superposition and entanglement. These inherent quantum properties streamline enhanced feature representation and the exploration of complex data structures, offering the potential to address challenges such as imbalanced datasets \cite{tudiscoevaluating2024}, a prevalent issue in healthcare applications.


This work introduces a novel approach to improve the classification performance of \textbf{Quantum Neural Networks} (\textbf{QNNs}) \cite{Abbas2021-yv} consisting of multiple \textbf{Variational Quantum Circuits} (\textbf{VQCs}) \cite{schuld2020circuit}, i.e. a parameterized quantum circuit optimized through iterative training, arranged sequentially. The proposed method enhances the model's nonlinearity by leveraging the measurement operation, and so its ability to detect and classify complex patterns.

The performance of this strategy is evaluated against classical machine learning models --- including Logistic Regression, Decision Tree, Random Forest, and Support Vector Machine (SVM) --- while varying its degrees of freedom, specifically the number of VQCs, across three widely-used healthcare datasets: Heart Failure \cite{heart_failure_clinical_records_519}, and Diabetes \cite{diabetes_dataset}, and Prostate Cancer \cite{prostate_cancer}. These datasets were chosen to represent diverse diagnostic challenges.

To address the constraints of current Noisy Intermediate-Scale Quantum (NISQ) devices and quantum emulators, a \textbf{Principal Component Analysis} (\textbf{PCA}) \cite{PCA} preprocessing step was applied to reduce the dimensionality of the datasets. This preprocessing step ensured compatibility with quantum hardware limitations while retaining the most informative features, enhancing the scalability and overall performance of the quantum classifiers.


The experimental results highlight the potential of the quantum model and validate the proposed approach, demonstrating that its advantages become more evident with increasing classification complexity. Specifically, the quantum models consistently outperform classical counterparts across all the datasets considered, with their advantage growing as dataset imbalance increases. Moreover, the proposed Multi-VQC approach proves to be particularly effective when classification becomes more challenging as the number of principal components retained after PCA decreases or when the dataset imbalance is less prominent.

Even though preliminary, these outcomes underscore the promise of quantum models for healthcare classification tasks. They emphasize the need for further research into hybrid quantum-classical approaches, which could open the way for the practical adoption of quantum algorithms in real-world medical applications, finally enhancing diagnostic accuracy and reliability.

The rest of the article is organized as follows. 
Section~\ref{sec:MotivationGen} outlines the motivation behind this work and the general idea, while Section~\ref{sec:Implementation} presents the actual models implementation and their characteristics. The obtained results are presented and discussed in Section~\ref{sec:Results}. Finally, in Section~\ref{sec:conclusions}, conclusions are drawn, and future perspectives are illustrated.


\section{Motivations and general idea} \label{sec:MotivationGen}

Healthcare diagnostic classification tasks often involve \textbf{imbalanced datasets}, where positive cases (e.g., diseased patients) are significantly exceeded by negative ones (e.g., healthy individuals) or vice versa. This imbalance presents a major challenge for ML models, as it can lead to biased predictions and poor generalization. Quantum models offer a promising alternative, as they exploit \textbf{higher-dimensional Hilbert spaces}, enabled by quantum phenomena such as superposition and entanglement, to improve classification performance. However, the practical deployment of quantum models on current \textbf{Noisy Intermediate-Scale Quantum (NISQ)} devices and quantum emulators is constrained by hardware limitations, particularly the limited number of handleable qubits. These restrictions lead to the employment of dimensionality reduction techniques such as PCA, which further complicates the classification task by requiring models to extract meaningful patterns from reduced feature spaces.

This work investigates the potential of quantum classifiers in healthcare by focusing on a popular near-term paradigm, \textbf{Quantum Neural Networks (QNNs)}, and analyzing the impact of their hyperparameters on disease detection accuracy. Furthermore, a novel technique, called \textbf{Multi-VQC}, is introduced. This method consists of multiple \textbf{Variational Quantum Circuits (VQCs)} --- parameterized quantum circuits optimized through iterative training --- arranged sequentially. The \textbf{Multi-VQC} approach aims to enhance the model’s expressivity by exploiting measurement operations to increase non-linearity, improving its ability to capture complex patterns within the data.

To evaluate the effectiveness of quantum models, their performance is compared against four classical machine learning algorithms: \textbf{Logistic Regression (LR), Decision Tree (DT), Random Forest (RF), and Support Vector Machine (SVM)}. The evaluation is conducted on three distinct health datasets --- \textbf{Prostate Cancer, Heart Failure, and Diabetes} --- selected for their limited number of features and elements (a necessary condition considering the computational constraints of quantum circuit simulations) and the inherent imbalance of classes, to evaluate the performance of the models in a real-world scenario.



\begin{figure}[h]
    \centering \vspace{-10pt}
    \includegraphics[width=0.7\linewidth]{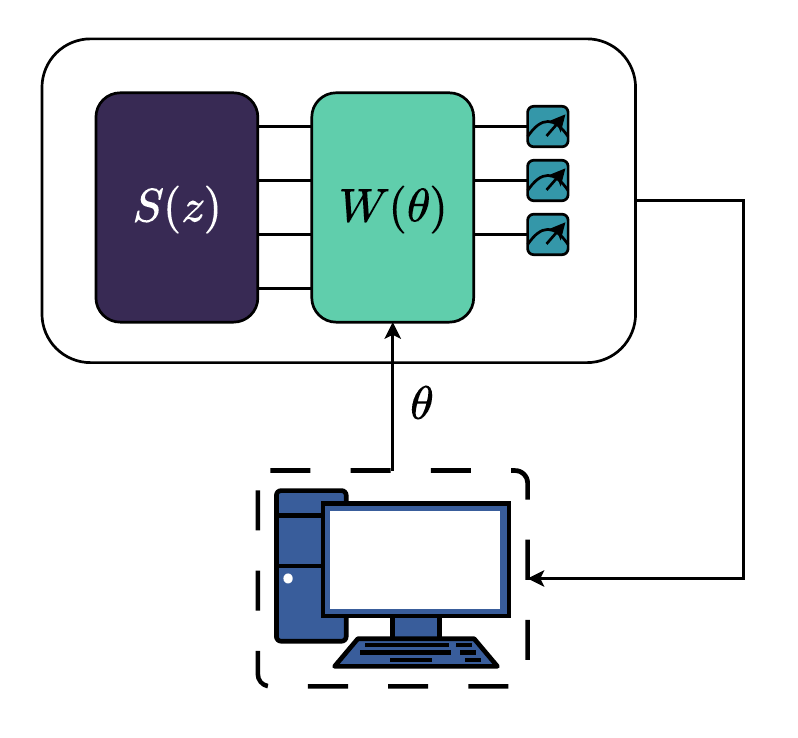}
    \vspace{-10pt}
    \caption{Representation of a Variational Quantum Circuit. \vspace{-18pt}}
    \label{fig:vqc}
\end{figure}



\vspace{-5pt}
\section{Implementation} \label{sec:Implementation}
\begin{figure}[h]
    \centering \vspace{-10pt}
    \includegraphics[width=\linewidth]{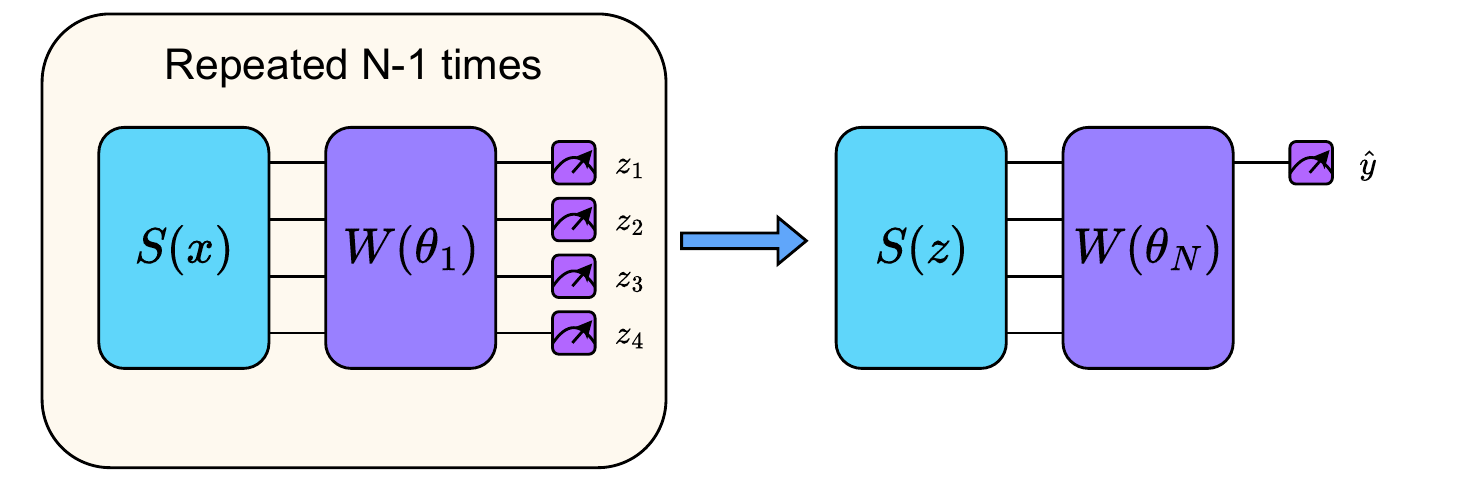}
    \caption{Representation of a Multi-VQC model. \vspace{-10pt}}
    \label{fig:multivqc}
\end{figure}

\vspace{-3pt}
This article presents a new technique called Multi-VQC for QNNs. In this approach, several Variational Quantum Circuits (VQCs) are placed sequentially, as shown in Figure \ref{fig:multivqc}, such that the output of the $i^{\textrm{th}}$ VQC serves as the input for the $(i+1)^{\textrm{th}}$ one. In all VQCs except the final one, each qubit is measured with the Pauli-Z observable, providing a value in the interval $[-1, 1]$ based on the probability of measuring the state $\ket{0}$ on that specific qubit. On the other hand, in the final VQC the number of measured qubits corresponds to the number of classes in the dataset. The advantage of this technique is that the repeated use of the measurement operator introduces additional nonlinearity into the model.

\begin{figure}[h]
    \centering
    \includegraphics[width=0.9\linewidth]{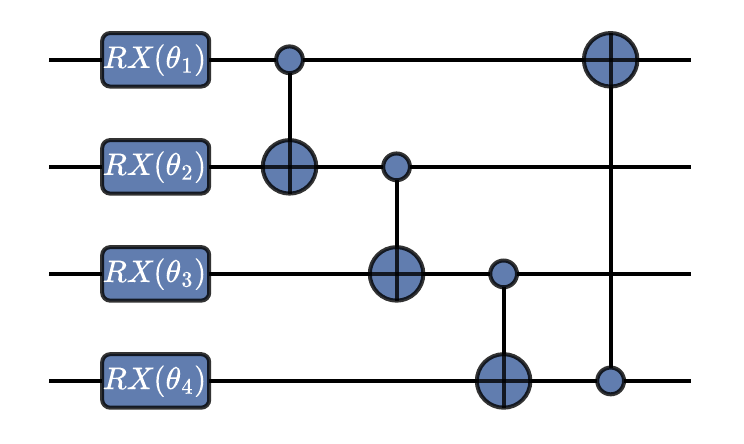}  \vspace{-10pt}
    \caption{Basic Entangling Layer.  \vspace{-15pt}}
    \label{fig:basic-ansatz}
\end{figure}

\begin{figure}[h]
    \centering
    \includegraphics[width=0.9\linewidth]{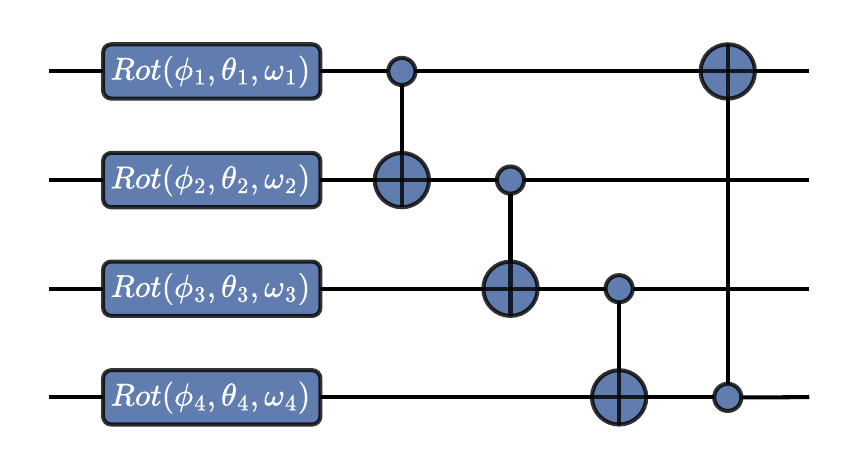}  \vspace{-10pt}
    \caption{Strongly Entangling Layer. \vspace{-15pt}}
    \label{fig:strongly-ansatz}
\end{figure}

The hyperparameters optimized to improve the model's effectiveness are: 
\begin{itemize}
    \item Encoding circuit.
    \item The ansatz topology.
    \item The reuploading technique.
    \item The number of layers.
\end{itemize}
Specifically, Angle Encoding was employed, with both RX and RY gates evaluated. Regarding the ansatz, two configurations were analyzed  \cite{bergholm2018pennylane}: Basic Entangling Layers, shown in Figure \ref{fig:basic-ansatz}, which use only RX gates, and Strongly Entangling Layers, represented in Figure \ref{fig:strongly-ansatz}, which implement a sequence of RZ-RY-RZ gates. Additionally, the impact of reuploading \cite{kolle2023weight} techniques was evaluated. Finally, the optimal number of layers was determined using a sort of early-stopping approach, where the layer count was incrementally increased until the validation loss failed to improve for $N$ consecutive steps. In this work, $N$ is set equal to the number of qubits.

The proposed Multi-VQC technique was evaluated by progressively increasing the number of VQCs up to three, while independently optimizing the number of layers for each configuration.
\begin{figure*}[t]
	\begin{subfigure}[t]{0.32\textwidth}
    \includegraphics[width=1\linewidth]{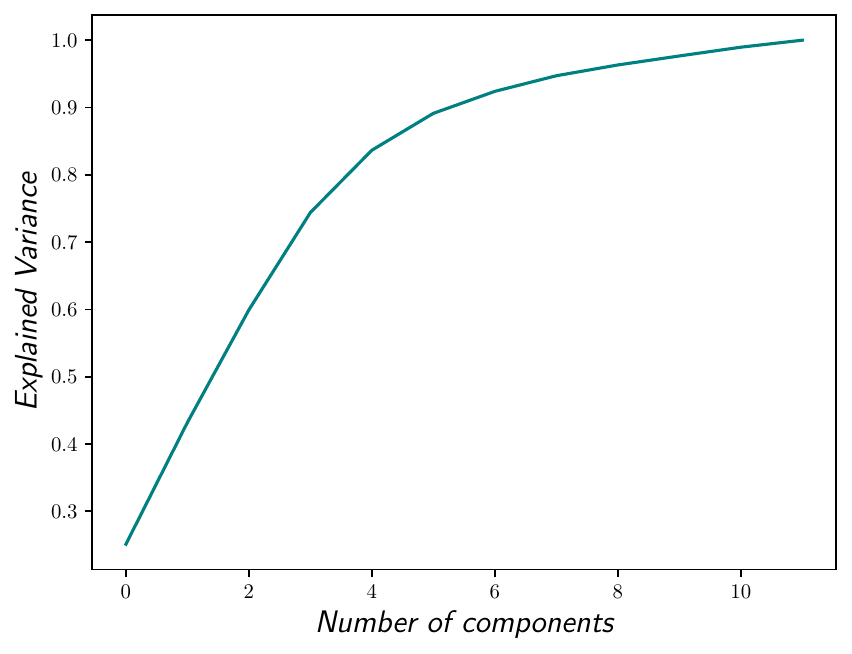} \vspace{-15pt}
    \caption{ Heart Failure dataset. Notice that 5 features include more than 90\% of the variance. \vspace{-13pt}}
    \label{fig:PCA_VarianceHeart}
	\end{subfigure}
    \quad
 	\begin{subfigure}[t]{0.32\textwidth}
	        \centering
    \includegraphics[width=1\linewidth]{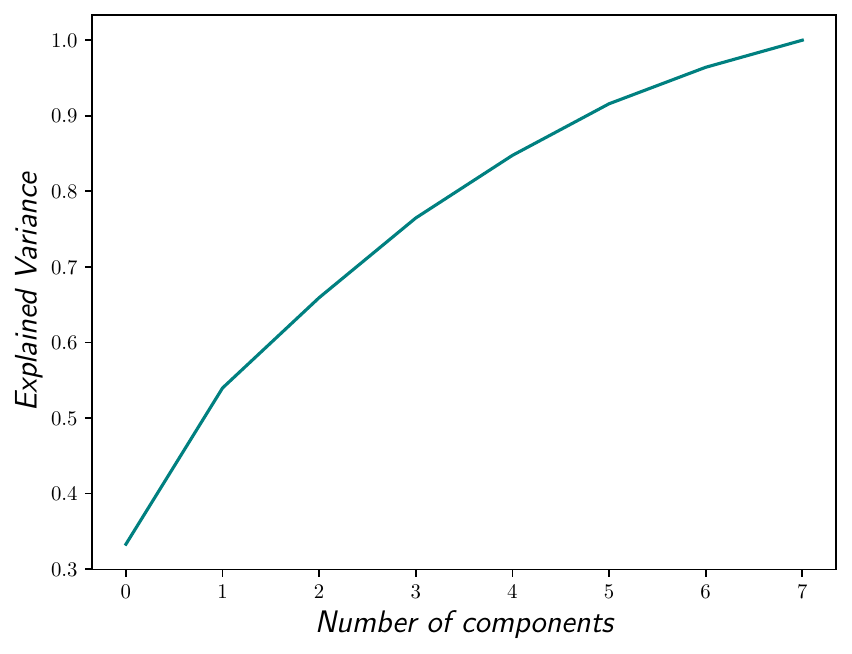}  \vspace{-15pt}
    \caption{Diabetes dataset. Notice that 6 features include more than 90\% of the variance. \vspace{-13pt} }
    \label{fig:PCA_VarianceDiabetes}
	\end{subfigure}
     \quad 
\begin{subfigure}[t]{0.32\textwidth}
    \includegraphics[width=1\linewidth]{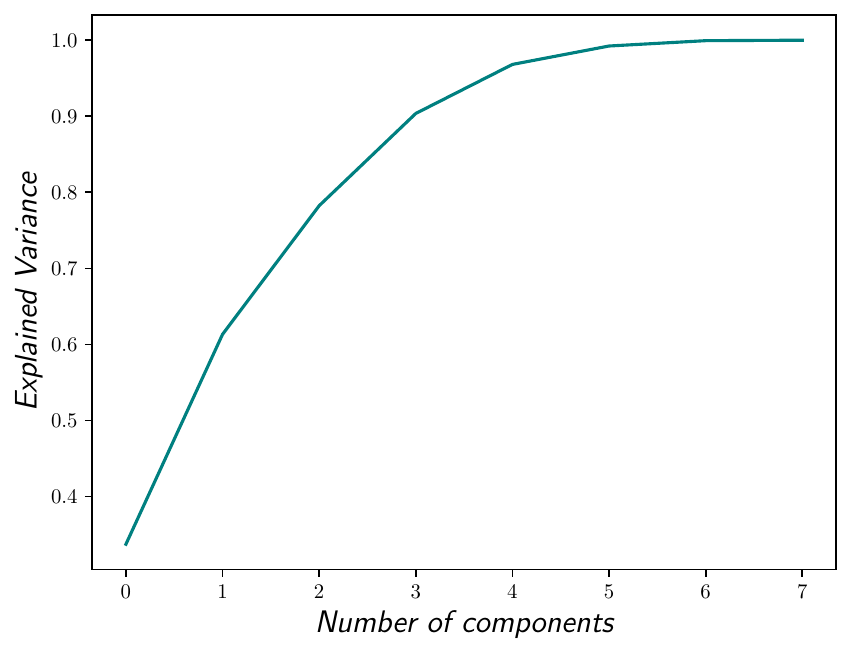}  \vspace{-15pt}
    \caption{Prostate Cancer dataset. Notice that 6 features include more than 99\% of the variance. \vspace{-13pt} }
    \label{fig:PCA_Varianceprostate}
	\end{subfigure}
	\caption{The cumulative explained variance for each component in the three considered datasets --- Heart Failure, Diabetes, and Prostate Cancer -- , applying Principal Component Analysis.  \vspace{-10pt} }
	\label{fig:PCA}
\end{figure*} 

The quantum models were compared with their classical counterparts, considering \textbf{Logistic Regression}, \textbf{Decision Tree}, \textbf{Random Forest}, and \textbf{Support Vector Machine} (\textbf{SVM}). In the case of the SVM, four kernel functions were considered: linear, polynomial (degree 3), radial basis function (RBF), and sigmoid.

All the quantum models are trained for 100 epochs, applying the early stopping technique with the patience parameter set to 5 to prevent overfitting and considering the Adam optimizer, which is particularly efficient.  Moreover,  a \textbf{balancing technique}  was applied to address the inherent class imbalance in healthcare datasets.  Specifically, a \textbf{weighting strategy} is employed to address class imbalance by assigning a higher weight to the minority class relative to the majority class. This is performed by scaling the weights: the weight of the minority class is multiplied by the proportion of elements in the majority class, and vice versa. This approach guarantees that the training process compensates for the imbalance, allowing the model to learn effectively from the minority class without being dominated by the majority class. Additionally, it eliminates the need for threshold optimization during classification, increasing the robustness of the prediction mechanism.

\vspace{-3pt}
\section{Results} \label{sec:Results}
\vspace{-3pt}

\subsection{Settings}
The employed models have been defined exploiting the PennyLane library (version 0.33), an open software framework for QML applications, which is compliant with classical machine learning libraries such as Tensorflow and Pytorch. In this work, the models are trained with Pytorch. For training and test the models, default.qubit simulator available in PennyLane was employed.

\subsection{Figures of Merit}

Healthcare datasets frequently exhibit significant class imbalance, making accuracy an incomplete figure of merit. Therefore, evaluation metrics that prioritize detecting minor class outcomes (e.g., cancer diagnoses) while minimizing false alarms are needed.

The first key metric is \textbf{Recall} (sensitivity), which measures a model's ability to identify all relevant positive cases:
\begin{equation}
\textrm{R} = \frac{\textrm{TP}}{\textrm{TP} +\textrm{FN}} \, ,
\label{eq:recall }
\end{equation}
where TP is the true positives count and FN indicates the false negatives. A high Recall guarantees the minimization of missed positive elements.

A second metric, \textbf{Precision} evaluates the portion of correctly identified positives among all instances predicted as positive:
\begin{equation}
\textrm{P} = \frac{\textrm{TP}}{\textrm{TP} +\textrm{FP}} \, ,
\label{eq:precision }
\end{equation}
where FP stands for the amount of false positives. The models characterized by high Precision reduce the risk of positive misprediction.

Finally, the \textbf{F1-score} balances these two metrics by combining  Precision and Recall via their harmonic mean:
\begin{equation}
\textrm{F1Score} = \frac{2 \cdot \textrm{R} \cdot \textrm{P}}{\textrm{R} + \textrm{P}} \, .
\label{eq:F1Score }
\end{equation}
F1-score is particularly useful in imbalanced healthcare contexts as it penalizes models that sacrifice either critical true positives (Recall) or specificity (Precision), ensuring robust performance for the minority class.

\subsection{Datasets}
The models are evaluated on three datasets: the \textbf{Heart Failure}\cite{heart_failure_clinical_records_519}, the \textbf{Diabetes}\cite{diabetes_dataset} and the \textbf{Prostate Cancer}\cite{prostate_cancer}. 

The Heart Failure dataset includes information from 299 patients collected over a follow-up period, described by 13 features. The label indicates whether the patient survived or died during the follow-up period. The 32\% of the elements belong to class 1, while the others belong to class 0.

The Diabetes dataset comprises 768 elements, each characterized by 8 features, while the output labels correspond to whether a patient has diabetes. The 34.9\% of elements belong to class 1, while the others belong to class 0.

Lastly, the Prostate Cancer dataset consists of data from 100 patients described by 9 variables. The two possible classes are M (malignant) and B (benign). The malignant elements correspond to 62\% of the elements in the entire data set, while the others are benign.

All datasets undergo an initial pre-processing stage. First, they are normalized using min-max scaling to ensure that features share a consistent scale and to prevent features with larger numerical ranges from dominating the training process. Next, PCA is applied to reduce dimensionality by identifying the directions of greatest variance, thereby retaining essential information and mitigating noise or redundancies. Finally, after dimensionality reduction, a second normalization step is performed to meet the specific requirements of the quantum circuit, ensuring the classical data embedding.

\subsection{Performed Tests}
 \subsubsection{Heart-Failure Dataset}

  {\renewcommand{\arraystretch}{1.2}
\begin{table}[h]
 \centering
 \caption{Precision (P) and Recall (R) obtained by QNN on the Heart Failure dataset.}
 \begin{widetable}{\columnwidth}{cccccccccccc}
 \hline
 \multirow{2}{*}{\textbf{Feat}} & \multirow{2}{*}{\textbf{\# VQC}} & \multirow{2}{*}{\textbf{Enc}} &  \multirow{2}{*}{\textbf{Reup}} & \multirow{2}{*}{\textbf{Ansatz}} & \multirow{2}{*}{\textbf{Layers}} & \multicolumn{2}{c}{\textbf{Train}} & \multicolumn{2}{c}{\textbf{Validation}} & \multicolumn{2}{c}{\textbf{Test}}\\
 & & & & & & \textbf{P} & \textbf{R} & \textbf{P} & \textbf{R} & \textbf{P} & \textbf{R} \\
 \hline
2 & 1 & X & True & basic & 7 & 0.34 & 0.73 & 0.39 & 1.00 & 0.33 & 0.84 \\
2 & 2 & Y & True & strongly & 4 & 0.49 & 0.58 & 0.47 & 0.60 & 0.32 & 0.32 \\
2 & 3 & X & True & basic & 6 & 0.39 & 0.69 & 0.50 & 0.80 & 0.29 & 0.53 \\ \hline
3 & 1 & X & False & basic & 3 & 0.40 & 0.52 & 0.60 & 0.60 & 0.22 & 0.26 \\
3 & 2 & Y & True & strongly & 14 & 0.63 & 0.71 & 0.64 & 0.60 & 0.35 & 0.32 \\
3 & 3 & Y & True & basic & 5 & 0.44 & 0.66 & 0.57 & 0.80 & 0.21 & 0.32 \\ \hline
4 & 1 & X & True & basic & 7 & 0.46 & 0.65 & 0.57 & 0.80 & 0.32 & 0.47 \\
4 & 2 & X & True & strongly & 7 & 0.52 & 0.69 & 0.55 & 0.73 & 0.29 & 0.26 \\
4 & 3 & Y & True & strongly & 11 & 0.59 & 0.74 & 0.52 & 0.80 & 0.29 & 0.26 \\ \hline
5 & 1 & X & False & basic & 2 & 0.41 & 0.63 & 0.56 & 0.67 & 0.30 & 0.53 \\
5 & 2 & Y & True & basic & 34 & 0.62 & 0.82 & 0.59 & 0.87 & 0.26 & 0.32 \\
5 & 3 & Y & True & strongly & 15 & 0.62 & 0.82 & 0.65 & 0.87 & 0.38 & 0.32 \\
 \hline
 \end{widetable} \vspace{-10pt}
 \label{tab:heart-failureVQC}
 \end{table}
}

Table~\ref{tab:heart-failureVQC} presents the Precision and Recall achieved by the best-performing QNN models on the Heart Failure dataset, considering different numbers of VQCs and features. It can be observed that single-VQC QNN models generally achieve higher Recall on both the training and validation sets, particularly in lower-feature configurations. This is advantageous in applications where minimizing false negatives is critical. On the other hand, Multi-VQC QNN models exhibit a more balanced trade-off between Precision and Recall, particularly when fewer features are considered. However, these models tend to suffer from overfitting, as indicated by the significant discrepancy between validation and test set results. This is particularly evident in cases with higher-layer architectures, where validation metrics remain high, but test performance worsens. These findings suggest that while additional VQCs improve generalization during training, they may introduce excessive model complexity for this use case, reducing the robustness of unseen data.
\begin{figure}[h]
    \centering
    \includegraphics[width=0.7\linewidth]{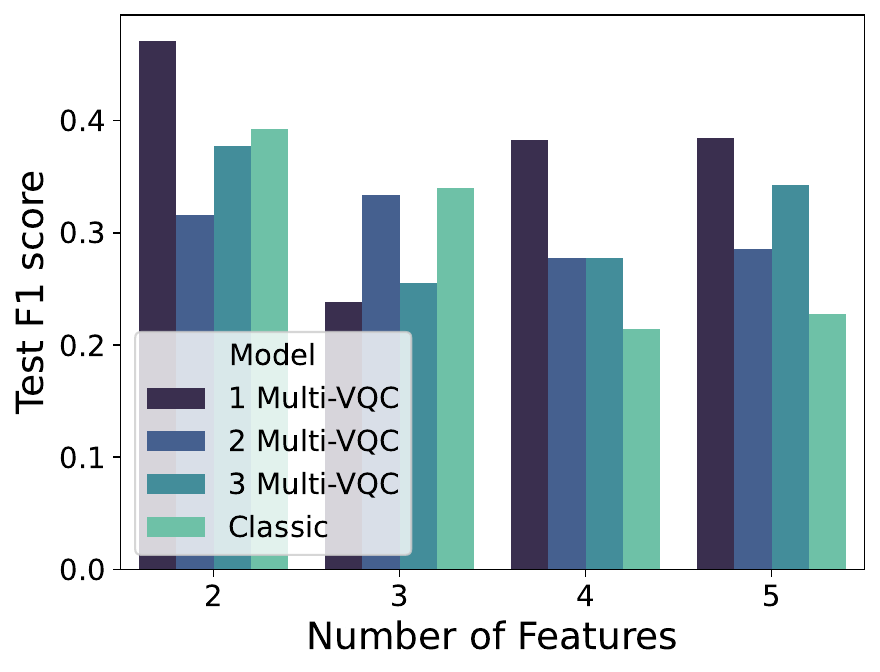}
    \caption{F1-score of the best models grouped by the number of VQCs on the Heart Failure test set. In this case, increasing the number of VQC does not guarantee improvements in the performance of the model, except for the dataset reduced to 3 features.}
    \label{fig:Heart-failure_results}
\end{figure} 

The quantum models are also compared in terms of F1-score for Heart-failure dataset in Figure \ref{fig:Heart-failure_results}. It shows that, except for the three features case, the single VQC performs better than Multi-VQC models due to overfitting. 

{\renewcommand{\arraystretch}{1.2}
\begin{table}[h]
 \centering
 \caption{Precision (P) and Recall (R) obtained by classical models on the Heart Failure dataset.}
 \begin{widetable}{\columnwidth}{ccccccccc}
 \hline
 \multirow{2}{*}{\textbf{Feat}} & \multirow{2}{*}{\textbf{Model}} & \multirow{2}{*}{\textbf{Kernel}} & \multicolumn{2}{c}{\textbf{Train}} & \multicolumn{2}{c}{\textbf{Validation}} & \multicolumn{2}{c}{\textbf{Test}}\\
 & & & \textbf{P} & \textbf{R} & \textbf{P} & \textbf{R} & \textbf{P} & \textbf{R} \\
 \hline
2 & LogisticRegression & - & 0.35 & 0.55 & 0.47 & 0.60 & 0.30 & 0.58 \\
3 & svm & rbf & 0.43 & 0.66 & 0.47 & 0.60 & 0.26 & 0.47 \\
4 & DecisionTree & - & 1.00 & 1.00 & 0.54 & 0.47 & 0.33 & 0.16 \\
5 & svm & poly & 0.46 & 0.71 & 0.50 & 0.80 & 0.20 & 0.26 \\
 \hline
 \end{widetable} \vspace{-10pt}
 \label{tab:heart-failure}
 \end{table}
}

{\renewcommand{\arraystretch}{1.2}
\begin{table}[h] 
 \centering
 \caption{Comparison of classical and quantum models in terms of Precision (P) and Recall (R) on the test set for Heart Failure dataset.} 
 \begin{widetable}{\columnwidth}{ccccccc}
 \hline
 \multirow{2}{*}{\textbf{Feat}} & \multicolumn{2}{c}{\textbf{QNN}} & \multicolumn{2}{c}{\textbf{QNN Multi-VQC}} & \multicolumn{2}{c}{\textbf{Classical}}\\
 & \textbf{P} & \textbf{R} & \textbf{P} & \textbf{R} & \textbf{P} & \textbf{R} \\
 \hline
2 &\textbf{0.33} & \textbf{0.84} & 0.29 & 0.53 & 0.30 & 0.58 \\
3 &0.22 & 0.26 & \textbf{0.35} & 0.32 & 0.26 & \textbf{0.47} \\
4 & 0.32 & \textbf{0.47} & 0.29 & 0.26 & \textbf{0.33} & 0.16 \\
5 &0.30 & 0.53 &\textbf{ 0.38 }&\textbf{ 0.32} & 0.20 & 0.26 \\
 \hline
 \end{widetable}  \vspace{-10pt}
 \label{tab:comparisonHeart}
 \end{table} 
} 

Moreover, the quantum models are also compared with the best classical ones, whose results and configuration are reported in Table \ref{tab:heart-failure}. As the number of features increases, the classical models exhibit overfitting, achieving strong performance on the training and validation sets but performing poorly on the test set.  As shown in Table \ref{tab:comparisonHeart} and in Figure \ref{fig:Heart-failure_results}, the classical models achieve precision on the test set comparable to that of QNN. However, their recall declines significantly as the number of features increases, highlighting challenges in maintaining sensitivity with larger feature sets and the advantage of quantum models, which demonstrate greater robustness in recall across different feature configurations.


{\renewcommand{\arraystretch}{1.2}
\begin{table}[h]
 \centering
 \caption{Precision (P) and Recall (R) obtained by QNN on the Diabetes dataset.}
 \begin{widetable}{\columnwidth}{cccccccccccc}
 \hline
 \multirow{2}{*}{\textbf{Feat}} & \multirow{2}{*}{\textbf{\# VQC}} & \multirow{2}{*}{\textbf{Enc}} &  \multirow{2}{*}{\textbf{Reup}} & \multirow{2}{*}{\textbf{Ansatz}} & \multirow{2}{*}{\textbf{Layers}} & \multicolumn{2}{c}{\textbf{Train}} & \multicolumn{2}{c}{\textbf{Validation}} & \multicolumn{2}{c}{\textbf{Test}}\\
 & & & & & & \textbf{P} & \textbf{R} & \textbf{P} & \textbf{R} & \textbf{P} & \textbf{R} \\
 \hline
2 & 1 & X & True & strongly & 2 & 0.54 & 0.57 & 0.60 & 0.56 & 0.45 & 0.43 \\
2 & 2 & X & True & basic & 3 & 0.55 & 0.77 & 0.57 & 0.70 & 0.48 & 0.61 \\
2 & 3 & Y & True & strongly & 9 & 0.54 & 0.80 & 0.53 & 0.67 & 0.43 & 0.61 \\ \hline
3 & 1 & Y & True & basic & 11 & 0.55 & 0.70 & 0.57 & 0.77 & 0.55 & 0.69 \\
3 & 2 & X & True & strongly & 11 & 0.58 & 0.78 & 0.58 & 0.86 & 0.59 & 0.81 \\
3 & 3 & X & True & strongly & 7 & 0.57 & 0.79 & 0.57 & 0.86 & 0.63 & 0.83 \\ \hline
4 & 1 & Y & True & basic & 11 & 0.58 & 0.76 & 0.60 & 0.79 & 0.56 & 0.74 \\
4 & 2 & X & True & strongly & 9 & 0.59 & 0.79 & 0.59 & 0.84 & 0.58 & 0.78 \\
4 & 3 & Y & True & strongly & 10 & 0.61 & 0.85 & 0.58 & 0.84 & 0.59 & 0.81 \\ \hline
5 & 1 & Y & True & strongly & 6 & 0.56 & 0.70 & 0.72 & 0.72 & 0.55 & 0.59 \\
5 & 2 & Y & True & basic & 13 & 0.55 & 0.78 & 0.67 & 0.84 & 0.56 & 0.70 \\
5 & 3 & Y & True & strongly & 3 & 0.53 & 0.68 & 0.71 & 0.79 & 0.47 & 0.50 \\ \hline
6 & 1 & Y & True & basic & 18 & 0.58 & 0.77 & 0.54 & 0.70 & 0.69 & 0.78 \\
6 & 2 & X & True & strongly & 21 & 0.71 & 0.90 & 0.56 & 0.74 & 0.58 & 0.72 \\
6 & 3 & Y & True & basic & 21 & 0.65 & 0.88 & 0.57 & 0.72 & 0.56 & 0.83 \\
 \hline
 \end{widetable}
 \label{tab:diabetesVQC}
 \end{table}
}
\subsubsection{Diabetes Dataset}

Table \ref{tab:diabetesVQC} presents the results achieved in terms of Precision and Recall by the best-performing QNN models on the Diabetes dataset, by varying numbers of  numbers of VQCs and features.  In general, QNN achieve their best test performance with 6 features, indicating this as an optimal feature set for the Diabetes dataset under the given configurations. This is reasonable with expectation since, as observed from Figure \ref{fig:PCA_VarianceDiabetes},  each component introduces significant variability, making configurations with fewer features less effective. Furthermore, the ideal configuration of VQCs varies depending on the number of retained components:
\begin{itemize}
    \item 3-VQC architectures achieve optimal performance with 3 or 4 retained components;
    \item 2-VQC configurations yield the best results for 5 retained components;
    \item 1-VQC systems dominate when six components are preserved.
\end{itemize}

\begin{figure}[h]
    \centering
    \includegraphics[width=0.7\linewidth]{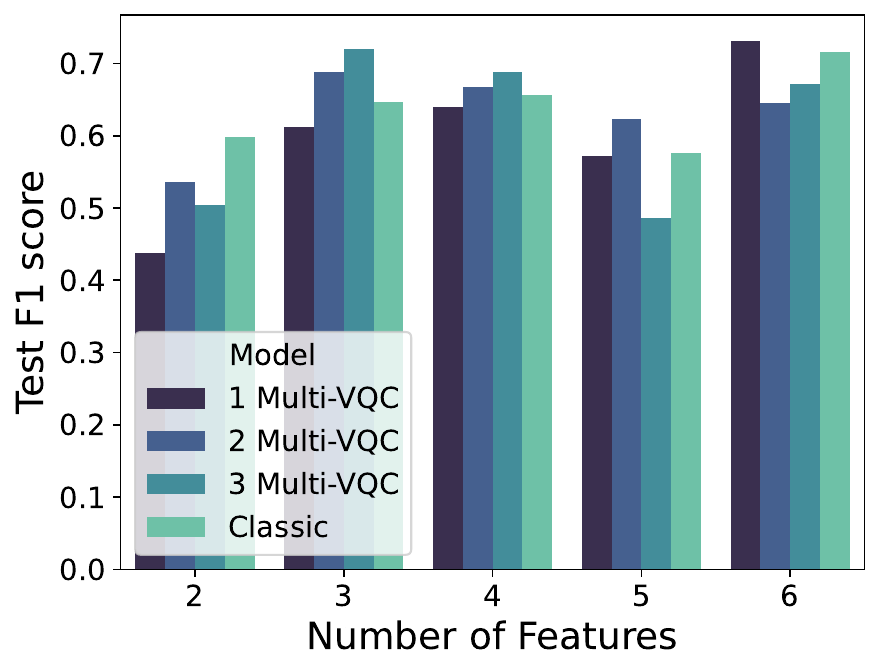}
    \caption{ F1-score of the best models grouped by the number of VQC on the diabetes test set. In this case, for a lower number of components, the Multi-VQC works better than the single VQC model. }
    \label{fig:diabetes_results}
\end{figure}

The quantum models are also evaluated in terms of F1-score for the Diabetes dataset in Figure \ref{fig:diabetes_results}. 

{\renewcommand{\arraystretch}{1.2}
\begin{table}[h]
 \centering
 \caption{Precision (P) and Recall (R) obtained by classical models on the Diabetes dataset.}
 \begin{widetable}{\columnwidth}{ccccccccc}
 \hline
 \multirow{2}{*}{\textbf{Feat}} & \multirow{2}{*}{\textbf{Model}} & \multirow{2}{*}{\textbf{Kernel}} & \multicolumn{2}{c}{\textbf{Train}} & \multicolumn{2}{c}{\textbf{Validation}} & \multicolumn{2}{c}{\textbf{Test}}\\
 & & & \textbf{P} & \textbf{R} & \textbf{P} & \textbf{R} & \textbf{P} & \textbf{R} \\
 \hline
2 & svm & rbf & 0.57 & 0.80 & 0.53 & 0.60 & 0.52 & 0.70 \\
3 & svm & rbf & 0.56 & 0.78 & 0.54 & 0.81 & 0.55 & 0.78 \\
4 & svm & rbf & 0.59 & 0.81 & 0.58 & 0.81 & 0.58 & 0.76 \\
5 & svm & linear & 0.57 & 0.68 & 0.72 & 0.72 & 0.53 & 0.63 \\
6 & svm & rbf & 0.62 & 0.87 & 0.55 & 0.77 & 0.64 & 0.81 \\
 \hline
 \end{widetable}
 \label{tab:diabetes}
 \end{table}
}

{\renewcommand{\arraystretch}{1.2}
\begin{table}[h] \vspace{-10pt}
 \centering
 \caption{Comparison of classical and quantum models in terms of Precision (P) and Recall (R) on test set for Diabetes dataset.}
 \begin{widetable}{\columnwidth}{ccccccc}
 \hline
 \multirow{2}{*}{\textbf{Feat}} & \multicolumn{2}{c}{\textbf{QNN}} & \multicolumn{2}{c}{\textbf{QNN Multi-VQC}} & \multicolumn{2}{c}{\textbf{Classical}}\\
 & \textbf{P} & \textbf{R} & \textbf{P} & \textbf{R} & \textbf{P} & \textbf{R} \\
 \hline
2 &0.45 & 0.43 & 0.48 & 0.61 &\textbf{ 0.52} & \textbf{0.70} \\
3 &0.55 & 0.69 & \textbf{0.63} & \textbf{0.83} & 0.55 & 0.78 \\
4 & 0.56 & 0.74 & \textbf{0.59} & \textbf{0.81} & 0.58 & 0.76 \\
5 &0.55 & 0.59  & \textbf{0.56} & \textbf{0.70} & 0.53 & 0.63 \\
6 &\textbf{0.69} & 0.78 & 0.56 & \textbf{0.83} & 0.64 & 0.81 \\
 \hline
 \end{widetable}
 \label{tab:comparisonDiabetes}
 \end{table}
}

The best classical models among those considered were also identified and reported in Table \ref{tab:diabetes}. For this dataset, the best-performing classical model is almost always the SVM with the radial basis function. In some cases, such as with five features, overfitting can be observed as the Precision and Recall in validation are significantly higher than those in the test set. As shown in Table \ref{tab:comparisonDiabetes}, the classical models achieve the best performance for 2 features. On the other hand, Multi-VQC QNN gives the best performance in all the other cases. These conclusions can also be drawn by comparing quantum and classical models in terms of F1-score, as shown in Figure \ref{fig:diabetes_results}.

Therefore, less complex quantum models become increasingly advantageous as the feature space expands, suggesting a balance between the complexity of the model and the type of classification faced.

{\renewcommand{\arraystretch}{1.2}
\begin{table}
 \centering
 \caption{Precision (P) and Recall (R) obtained by QNN on the Prostate Cancer dataset.}
 \begin{widetable}{\columnwidth}{cccccccccccc}
 \hline
 \multirow{2}{*}{\textbf{Feat}} & \multirow{2}{*}{\textbf{\#VQC}} & \multirow{2}{*}{\textbf{Enc}} &  \multirow{2}{*}{\textbf{Reup}} & \multirow{2}{*}{\textbf{Ansatz}} & \multirow{2}{*}{\textbf{Layers}} & \multicolumn{2}{c}{\textbf{Train}} & \multicolumn{2}{c}{\textbf{Validation}} & \multicolumn{2}{c}{\textbf{Test}}\\
 & & & & & & \textbf{P} & \textbf{R} & \textbf{P} & \textbf{R} & \textbf{P} & \textbf{R} \\
 \hline
2 & 1 & Y & False & basic & 2 & 0.64 & 0.88 & 0.71 & 1.00 & 0.73 & 0.92 \\
2 & 2 & Y & False & basic & 2 & 0.70 & 0.95 & 0.71 & 1.00 & 0.75 & 1.00 \\
2 & 3 & Y & True & strongly & 8 & 0.80 & 0.93 & 0.80 & 0.80 & 0.92 & 1.00 \\ \hline
3 & 1 & Y & True & basic & 4 & 0.70 & 0.88 & 0.75 & 0.90 & 0.80 & 1.00 \\
3 & 2 & X & False & basic & 4 & 0.77 & 0.93 & 0.77 & 1.00 & 0.92 & 1.00 \\
3 & 3 & X & True & strongly & 11 & 0.89 & 0.97 & 0.77 & 1.00 & 0.80 & 1.00 \\ \hline
4 & 1 & X & False & basic & 14 & 0.74 & 0.85 & 0.83 & 1.00 & 1.00 & 0.83 \\
4 & 2 & X & False & strongly & 3 & 0.79 & 0.93 & 0.83 & 1.00 & 1.00 & 0.83 \\
4 & 3 & Y & False & strongly & 5 & 0.67 & 0.97 & 0.83 & 1.00 & 0.75 & 1.00 \\ \hline
5 & 1 & Y & False & basic & 4 & 0.74 & 0.88 & 0.90 & 0.90 & 0.80 & 1.00 \\
5 & 2 & X & False & basic & 12 & 0.72 & 0.72 & 0.90 & 0.90 & 0.67 & 0.50 \\
5 & 3 & X & True & strongly & 13 & 1.00 & 1.00 & 0.83 & 1.00 & 0.73 & 0.67 \\ \hline
6 & 1 & X & False & basic & 10 & 0.77 & 0.60 & 0.82 & 0.90 & 0.67 & 0.50 \\
6 & 2 & X & False & strongly & 8 & 0.88 & 0.90 & 0.90 & 0.90 & 1.00 & 0.83 \\
6 & 3 & X & True & strongly & 9 & 0.94 & 0.82 & 0.82 & 0.90 & 0.80 & 0.67 \\
 \hline
 \end{widetable}
 \label{tab:prostateVQC}
 \end{table}
}

\subsubsection{Prostate Cancer Dataset}

Table \ref{tab:prostateVQC} shows the results obtained in terms of Precision and Recall by the best-performing QNN models on the Prostate Cancer dataset, by varying numbers of VQCs and features considered.  The optimal number of VQC strongly depends on the feature count: 
\begin{itemize}
\item The 3-VQC architecture prevails when fewer PCA components are retained, mitigating challenges from information loss in dimensionality reduction.
\item The 2-VQC configuration dominates for larger feature sets, except with 5 components where the 1-VQC structure achieves better performance.
\end{itemize}

\begin{figure}[h]
    \centering
    \includegraphics[width=0.7\linewidth]{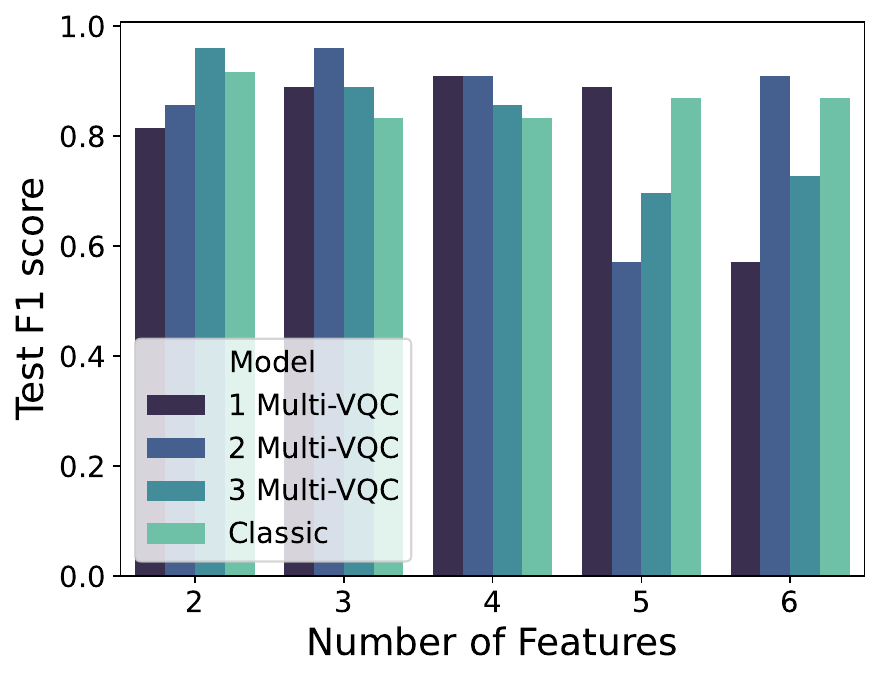} 
    \caption{F1-score of the best models grouped by the number of VQCs on the prostate cancer test set. The Multi-VQC approach has obtained better results than the single VQC, except for the case with 5 features.} 
    \label{fig:ProstateCancer_results}
\end{figure}

The quantum models are also compared in terms of F1-score for the Prostate Cancer dataset in Figure \ref{fig:ProstateCancer_results}. 

{\renewcommand{\arraystretch}{1.2}
\begin{table}
 \centering
 \caption{Precision (P) and Recall (R) obtained by classical models on the Prostate Cancer dataset.}
 \begin{widetable}{\columnwidth}{ccccccccc}
 \hline
 \multirow{2}{*}{\textbf{Feat}} & \multirow{2}{*}{\textbf{Model}} & \multirow{2}{*}{\textbf{Kernel}} & \multicolumn{2}{c}{\textbf{Train}} & \multicolumn{2}{c}{\textbf{Validation}} & \multicolumn{2}{c}{\textbf{Test}}\\
 & & & \textbf{P} & \textbf{R} & \textbf{P} & \textbf{R} & \textbf{P} & \textbf{R} \\
 \hline
2 & RandomForest & - & 1.00 & 1.00 & 0.78 & 0.70 & 0.92 & 0.92 \\
3 & RandomForest & - & 1.00 & 1.00 & 0.82 & 0.90 & 0.83 & 0.83 \\
4 & RandomForest & - & 1.00 & 1.00 & 0.82 & 0.90 & 0.83 & 0.83 \\
5 & DecisionTree & - & 1.00 & 1.00 & 0.80 & 0.80 & 0.91 & 0.83 \\
6 & RandomForest & - & 1.00 & 1.00 & 0.82 & 0.90 & 0.91 & 0.83 \\
 \hline
 \end{widetable} \vspace{-5pt}
 \label{tab:prostate}
 \end{table}
}

{\renewcommand{\arraystretch}{1.2}
\begin{table}[h] \vspace{-5pt}
 \centering
 \caption{Comparison of classical and quantum models in terms of Precision (P) and Recall (R) on the test set for Prostate Cancer dataset.}
 \begin{widetable}{\columnwidth}{ccccccc}
 \hline
 \multirow{2}{*}{\textbf{Feat}} & \multicolumn{2}{c}{\textbf{QNN}} & \multicolumn{2}{c}{\textbf{QNN Multi-VQC}} & \multicolumn{2}{c}{\textbf{Classical}}\\
 & \textbf{P} & \textbf{R} & \textbf{P} & \textbf{R} & \textbf{P} & \textbf{R} \\
 \hline
2 &0.73 & 0.92 & \textbf{0.92} & \textbf{1.00} & 0.92 & 0.92 \\
3 &0.80 & 1.00 & \textbf{0.92} & \textbf{1.00} & 0.83 & 0.83 \\
4 &\textbf{1.00} & \textbf{0.83} & \textbf{1.00} & \textbf{0.83} & 0.83 & 0.83 \\
5 &0.80 & \textbf{1.00} & 0.73 & 0.67  & \textbf{0.91} & 0.83 \\
6 &0.67 & 0.50 & \textbf{1.00} & \textbf{0.83} & 0.91 & 0.83 \\
 \hline
 \end{widetable} \vspace{-10pt}
 \label{tab:comparisonProstate}
 \end{table}
}

The best classical models among those considered were also identified and reported in Table \ref{tab:prostate}. Random Forest results in general as the best performing classical model among those considered in this work. As shown in Table \ref{tab:comparisonProstate}, quantum models outperform the classical ones in all the cases.  For this dataset, Multi-VQC QNN models perform exceptionally well in terms of both Precision and Recall.  Similar observations can also be done by comparing models in terms of F1-score, as shown in Figure \ref{fig:ProstateCancer_results}.


%
\subsection{Discussion}

As shown in Figures \ref{fig:Heart-failure_results}, \ref{fig:diabetes_results}, and \ref{fig:ProstateCancer_results}, quantum models consistently outperform classical models across the three healthcare classification tasks. However, the benefits of the Multi-VQC approach depend on dataset complexity and feature selection. 

From the results, Multi-VQC architectures show improved classification performance in some cases, particularly when the number of features is low. This suggests that sequential quantum circuits help capture complex patterns in reduced feature spaces. However, as the feature count increases, overfitting is observed, likely due to excessive parameterization without sufficient data to generalize.

This dimensional sensitivity suggests a practical workflow: initial implementation of a baseline 1-VQC model followed by incremental architectural expansion (2-VQC and 3-VQC) if classification performance is unsatisfactory.

\section{Conclusion} \label{sec:conclusions}
This article introduces a novel technique called Multi-VQC, which involves utilizing multiple VQCs in series to enhance classification performance by leveraging the non-linearity introduced by the measurement operation. The Multi-VQC approach is compared against the single-VQC method and classical Machine Learning models on three datasets --- Prostate Cancer, Heart Failure, and Diabetes --- characterized by class imbalance.

The results indicate that the Multi-VQC approach is particularly effective for complex classification tasks, while its advantages decrease as the classification task becomes simpler. This observation is supported by the performance of the 3-VQC, which consistently achieves superior F1-scores than the other model when retaining a low number of components, a scenario where classification is more challenging. This highlights the potential of the Multi-VQC technique in addressing challenging classification problems. Furthermore, this study demonstrates that quantum methods can achieve better results than their classical counterparts.

Future exploration could focus on experimenting with new ansatz topology and alternative encoding strategies. Additionally, validating the Multi-VQC technique on other datasets would further confirm its applicability and effectiveness.

\bibliographystyle{IEEEtran}
\bibliography{biblio}

\end{document}